\documentclass[sigconf]{acmart}
\AtBeginDocument{%
  }

\copyrightyear{2026}
\acmYear{2026}
\setcopyright{cc}
\setcctype{by}
\acmConference[DAC '26]{63rd ACM/IEEE Design Automation Conference}{July 26--29, 2026}{Long Beach, CA, USA}
\acmBooktitle{63rd ACM/IEEE Design Automation Conference (DAC '26), July 26--29, 2026, Long Beach, CA, USA}
\acmDOI{10.1145/3770743.3804218}
\acmISBN{979-8-4007-2254-7/2026/07}

\usepackage{booktabs}
\usepackage{multirow}
\usepackage{multicol}
\usepackage{tikz}
\begin{document}

\title{StepPRM-RTL: Stepwise Process-Reward Guided LLM Fine-Tuning for Enhanced RTL Synthesis}

\author{Prashanth Vijayaraghavan}
\email{prashanthv@ibm.com}
\affiliation{%
  \institution{IBM Research}
  \city{San Jose}
  \state{CA}
  \country{USA}
}

\author{Apoorva Nitsure}
\email{Apoorva.Nitsure@ibm.com}
\affiliation{%
  \institution{IBM Research}
  \city{San Jose}
  \state{CA}
  \country{USA}
}

\author{Luyao Shi}
\email{luyao.shi@ibm.com}
\affiliation{%
  \institution{IBM Research}
  \city{San Jose}
  \state{CA}
  \country{USA}
}

\author{Ehsan Degan}
\email{edehgha@us.ibm.com}
\affiliation{%
  \institution{IBM Research}
  \city{San Jose}
  \state{CA}
  \country{USA}
}
\author{Vandana Mukherjee}
\email{vandana@us.ibm.com}
\affiliation{%
  \institution{IBM Research}
  \city{San Jose}
  \state{CA}
  \country{USA}
}
\renewcommand{\shortauthors}{Vijayaraghavan et al.}

\begin{abstract}
  Automatic generation of RTL code for digital hardware designs remains challenging due to long-horizon reasoning, multi-step dependencies, and strict correctness constraints in Verilog and VHDL. We present StepPRM-RTL, a novel framework that combines stepwise trajectory modeling, process-reward modeling (PRM), and retrieval-augmented fine-tuning (RAFT) to enhance both the functional correctness and reasoning fidelity of LLM-based RTL code generation. StepPRM-RTL constructs stepwise reasoning trajectories from canonical solutions, where each step contains a rationale and incremental code modification. A Process Reward Model (PRM) evaluates intermediate steps, providing dense feedback that guides reinforcement-style updates during RAFT fine-tuning. Monte Carlo Tree Search (MCTS) explores alternative reasoning paths, enriching the training dataset with high-quality trajectories. This integration of stepwise and outcome-aware rewards allows the model to learn both how and why to construct correct RTL, improving long-horizon reasoning beyond standard supervised or outcome-based training. Experimental evaluation on benchmark Verilog and VHDL datasets demonstrates that StepPRM-RTL outperforms the best prior methods by over 10\% in functional correctness and reasoning fidelity metrics. Ablation studies confirm that the combination of PRM-guided rewards and stepwise trajectory exploration is key to its performance. StepPRM-RTL generalizes across RTL languages and provides a scalable framework for high-fidelity, interpretable code generation, establishing a new standard for LLM-assisted hardware design automation.
\end{abstract}


%%


\maketitle

\section{Introduction}

Automating Register–Transfer Level (RTL) code generation remains a central challenge in Electronic Design Automation (EDA). Unlike general-purpose programming, RTL demands not only syntactic correctness but also precise temporal, concurrent, and structural behaviors that govern circuit functionality. A single misaligned state update or improperly gated enable path can propagate across modules, breaking the datapath despite remaining syntactically valid. Consequently, generating semantically and functionally correct Verilog/VHDL code is both high-impact and underexplored, with immediate relevance to industrial design productivity.

Current RTL generation approaches~\cite{liu2023chipnemo, chipchat,chipformer,gpt4aigchip,vijayaraghavan2024chain}, primarily rely on supervised learning over code corpora, capturing surface-level patterns but not the reasoning sequence required to assemble correct control and datapath logic. Outcome-driven methods~\cite{WeiTanMendoza2025, akyash2025rtl++} evaluate correctness only at the final design level, offering no supervision for intermediate decisions, such as structuring reset logic, aligning control-path transitions, or coordinating enables across always blocks. As a result, these models struggle with long-horizon dependencies and cannot reliably shape multi-step design trajectories.

Recent advances in software code generation attempt to address these issues by introducing process reward models (PRMs) for intermediate-step scoring~\cite{LiDaiZhang2025, YeZhangJiang2025}. However, these PRMs operate at the token level, which is fundamentally mismatched to hardware semantics: meaningful RTL decisions often span statements, modules, and signal groups, making token-level credit assignment noisy and unstable. Moreover, structured search techniques such as Monte-Carlo Tree Search (MCTS), widely used in reasoning-intensive domains~\cite{kemmerling2024beyond,swiechowski2023monte}, remain largely unexplored for RTL synthesis.

To address these limitations, we propose \textbf{StepPRM-RTL}, a reasoning-aware RTL generation framework that introduces step-level supervision aligned with hardware semantics. Each reasoning step consists of an interpretable rationale paired with its corresponding code edit, enabling a Process Reward Model (StepPRM) to evaluate choices at the granularity of meaningful RTL behavior. StepPRM further supports \emph{PRM-guided MCTS exploration}, where the generator proposes alternative reasoning paths for the same specification, and MCTS evaluates them using step-level rewards and lightweight synthesizability checks. This produces a diverse set of high-value trajectories that extend beyond supervised decompositions while remaining grounded in verifiable hardware logic. Finally, StepPRM-RTL integrates these trajectories into a \emph{Retrieval-Augmented Fine-Tuning (RAFT)} framework~\cite{zhang2024raft}. RAFT retrieves canonical reasoning steps from similar designs and uses StepPRM-based intermediate rewards to stabilize policy refinement. This integrates step-level reasoning supervision, structured trajectory exploration, and retrieval-based context into a single coherent training pipeline, enabling effective long-horizon RTL code generation.
\begin{figure*}
    \centering
    \includegraphics[width=0.68\linewidth]{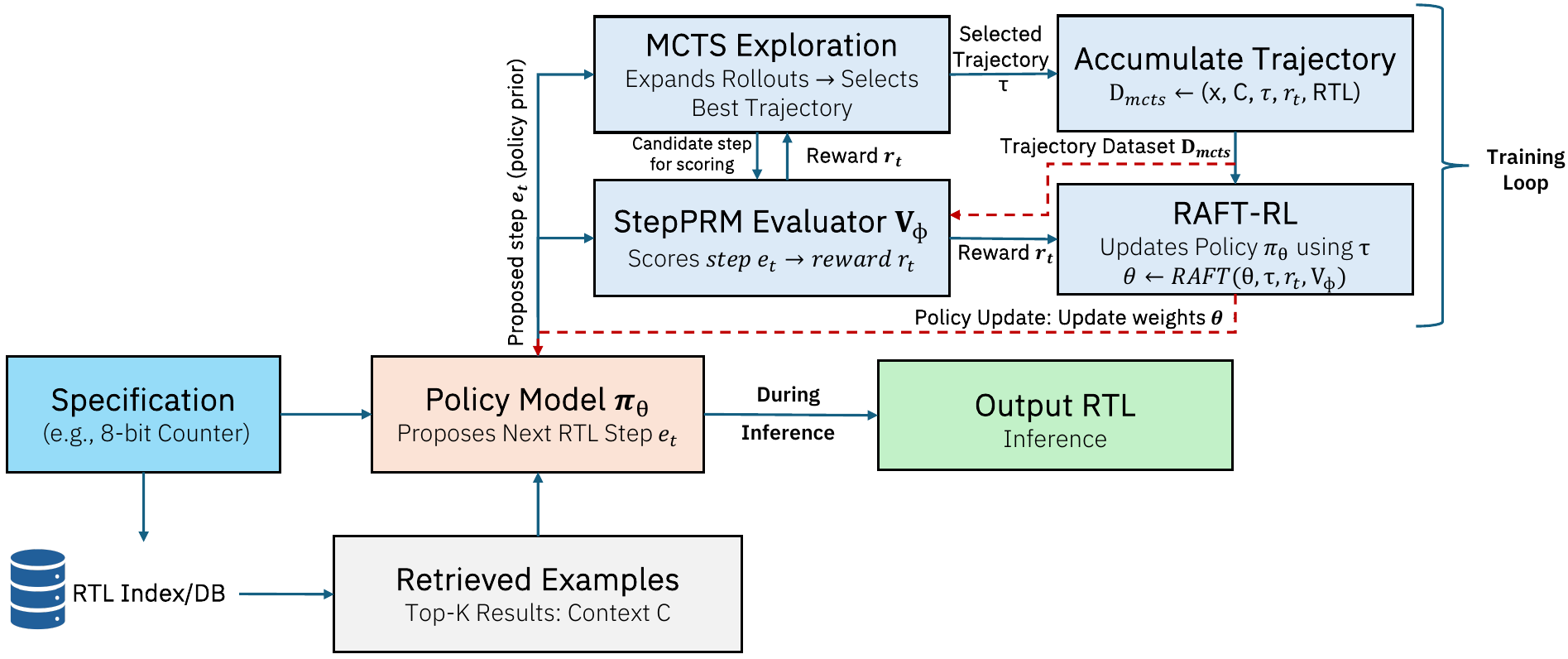}
    \caption{Overall Workflow of StepPRM-RTL: Training Loop (Top) and Inference (Center).}
    \label{fig:overview}
\end{figure*}

Figure~\ref{fig:overview} summarizes the workflow: (1) extract canonical stepwise trajectories; (2) expand the reasoning space using StepPRM-guided MCTS; (3) refine StepPRM on the expanded trajectory set; and (4) update the generator using RAFT with step-level rewards. This iterative loop jointly improves both the policy and the reward model while maintaining semantic alignment with RTL design principles. Our contributions are summarized as follows:

\noindent \textbf{Step-Level Process Rewarding for RTL:} We introduce \textbf{StepPRM-RTL}, the first framework to define and score semantically meaningful intermediate reasoning steps for HDL, resolving the mismatch between token-level scoring and hardware-level behaviors.

\noindent \textbf{Unified Reasoning Pipeline:} We propose an integrated pipeline combining StepPRM-guided MCTS exploration with RAFT-based policy refinement, enabling stable long-horizon credit assignment and retrieval-grounded reasoning.

\noindent \textbf{Comprehensive Evaluation:} Extensive experiments on Verilog and VHDL benchmarks demonstrate significant improvements in step-level reasoning quality, pass@k, functional correctness, and generalization over supervised and reward-based baselines.
% \end{enumerate}

\section{Problem Formulation}

We study the task of generating functionally correct Register–Transfer Level (RTL) designs from behavioral specifications. Let $x$ denote an input specification (e.g., a natural-language description of module behavior) and let $c^\star$ be a corresponding canonical Verilog/VHDL implementation. Instead of treating HDL generation as flat token prediction, we model RTL construction as a sequence of semantically meaningful \emph{design steps}. Each step is represented as $e_t=(r_t,\delta_t)$, where $r_t$ is a natural-language rationale describing a hardware design decision (such as adding synchronous reset or propagating enables), and $\delta_t$ is the code edit applied to the current partial implementation. Applying edits sequentially produces partial designs $c_0, c_1, \ldots, c_T$ with $c_t = \delta_t(c_{t-1})$, where $c_0$ is empty or templated and $c_T$ is the final design. A trajectory is thus $\tau=\langle e_1,\ldots,e_T\rangle$, generated by a policy model $\pi_\theta(e_t \mid x, c_{t-1})$ that conditions on both the specification and the evolving design state.

To supervise intermediate reasoning quality, we define a Step-level Process Reward Model (StepPRM) $V_\phi(e_t, c_{t-1}, x)$ that assigns a semantic score to each step, reflecting structural correctness, consistency with RTL design intent, and alignment with hardware semantics. Final correctness is assessed by an outcome reward $R_{\mathrm{out}}(c_T)$ based on compilation, simulation, and testbench verification. In this work, we leverage two types of datasets: an in-house RTL-IR corpus comprising combinations of spec, code and summary and derived stepwise trajectories, used to train both $\pi_\theta$ and $V_\phi$; and \textbf{Verilog-Eval} and \textbf{VHDL-Eval}, two strictly held-out benchmarks used only for evaluation of generalization and functional correctness.

\subsection{Objective}

Our goal is to learn a reasoning-aware policy that produces high-quality trajectories with sound intermediate decisions and correct final implementations. Formally, we maximize the expected trajectory value:
$
\label{eq:main_objective_inline}
\max_{\theta} \; 
\mathbb{E}_{\tau \sim \pi_\theta(\cdot \mid x)}
\left[ \mathcal{V}(\tau) \right],
$
where the value of a trajectory combines step-level and outcome-level rewards:
\begin{equation}
\mathcal{V}(\tau)
=
\alpha \sum_{t=1}^{T} V_\phi(e_t, c_{t-1}, x)
+
\beta \, R_{\mathrm{out}}(c_T),
\end{equation}
with $\alpha$ and $\beta$ weighting semantic reasoning quality and final functionality. This formulation provides dense, hardware-aligned supervision for long-horizon RTL construction.
\section{Methodology: StepPRM-RTL Framework}
\subsection{Overview}
We propose \textbf{StepPRM-RTL}\footnote{Short for Step-level Process Reward Model for RTL Synthesis}, an RL-guided framework for generating correct RTL designs from natural-language or structured specifications. StepPRM-RTL models RTL generation as a sequence of semantically meaningful design steps and integrates four tightly coupled components: (i) \emph{stepwise trajectory construction} from canonical RTL code to obtain high-quality reasoning demonstrations, (ii) a \emph{Step-level Process Reward Model (StepPRM)} $V_\phi$ that assigns semantic scores to intermediate design decisions, (iii) \emph{PRM-guided Monte-Carlo Tree Search (MCTS)} to explore alternative reasoning paths and collect diverse, high-value trajectories, and (iv) \emph{retrieval-augmented fine-tuning (RAFT)} that refines the generation policy $\pi_\theta$ using retrieved trajectories combined with StepPRM rewards. The framework operates as an iterative loop: canonical RTL implementations are first decomposed into stepwise trajectories to bootstrap the StepPRM. The initial PRM guides MCTS exploration to generate diverse, high-value reasoning trajectories beyond the canonical examples. These trajectories are then used to refine the PRM, improving its ability to assign semantically meaningful intermediate rewards. Finally, the generation policy $\pi_\theta$ is updated through RAFT fine-tuning, combining retrieved trajectory context with StepPRM scores to reinforce correct intermediate reasoning. This loop, consisting of trajectory collection, followed by PRM refinement, and then policy updates, repeats until convergence, ensuring continuous improvement of both the reward model and the generator. By unifying interpretable step supervision, structured exploration, and reward-guided policy refinement, StepPRM-RTL enhances long-horizon reasoning fidelity and final RTL correctness.

\subsection{Stepwise Trajectory Construction}

\begin{table*}[h!]
\centering
\begin{tabular}{|c|p{6cm}|p{9.5cm}|}
\hline
Step $e_t$ & Rationale $r_t$ & Code Edit $\delta_t$ \\
\hline
1 & Declare module interface and output registers & \texttt{module counter(clk, reset, out); input clk, reset; output reg [1:0] out;} \\
2 & Add always block triggered by clock and reset & \texttt{always @(posedge clk or posedge reset) begin ... end} \\
3 & Assign reset behavior to initialize counter & \texttt{if (reset) out <= 2'b00;} \\
4 & Increment counter on clock edge & \texttt{else out <= out + 1;} \\
5 & End module & \texttt{endmodule} \\
\hline
\end{tabular}
\caption{Stepwise decomposition of a 2-bit counter, pairing each rationale with its code edit.}
\label{tab:stepwise_trajectory}
\end{table*}

The first component of StepPRM-RTL is \emph{Stepwise Trajectory Construction}, which decomposes canonical RTL implementations into semantically meaningful intermediate design steps. Given a specification $x$ and its canonical RTL implementation $c^\star$ (Verilog or VHDL), we generate a stepwise trajectory, $\tau = \langle e_1, e_2, \dots, e_T \rangle$, where $e_t = (s_t, \delta_t)$, with each step $e_t$ pairing a human- or model-generated rationale statement $s_t$ and the corresponding code edit $\delta_t$ applied to a partial implementation $c_{t-1}$. The final partial implementation $c_T$ should reconstruct $c^\star$.

In practice, this decomposition leverages large language models (LLMs) to propose rationales and code edits, optionally assisted by abstract syntax tree (AST) analysis to ensure syntactic and structural consistency. This produces high-quality, interpretable demonstrations $(x, \tau)$, which provide supervised training pairs for initializing the generation policy $\pi_\theta$ and bootstrapping the Step-level Process Reward Model (StepPRM). Unlike token-level PRMs, StepPRM learns to assign semantic rewards to entire steps, improving stability and credit assignment in downstream reinforcement learning. Table~\ref{tab:stepwise_trajectory} illustrates the decomposition of a simple 2-bit counter design into a stepwise trajectory. Collectively, these trajectories form the initial training pool $\mathcal{D}_0$, used to initialize both StepPRM $V_\phi$ and the supervised policy $\pi_\theta$. This bootstrap phase establishes dense, step-level supervision that enables subsequent iterative exploration and reward-guided policy refinement.

\subsection{Step-Level Process Reward Model (StepPRM)}

Given the bootstrapped trajectory dataset $\mathcal{D}_0 = \{(x, \tau)\}$, the goal of the Step-Level Process Reward Model (StepPRM) is to learn a function $ V_{\phi} : (x, e_{1:t}) \mapsto \mathbb{R},$
which assigns a scalar reward to each intermediate step $e_t$ conditioned on the specification $x$ and the partial reasoning trajectory $e_{1:t} = \langle e_1, \dots, e_t \rangle$. Unlike token-level reward models, StepPRM operates at the \emph{semantic step} granularity, enabling stable credit assignment over long-horizon RTL synthesis trajectories.

\subsubsection{Supervised Preference Learning from Canonical Trajectories}
Each stepwise trajectory $\tau = \langle e_1, \dots, e_T \rangle$ obtained from canonical decomposition corresponds to a sequence of high-quality intermediate decisions. For StepPRM training, we treat each canonical step $e_t^\star$ as preferable to perturbed or low-quality steps $\tilde{e}_t$ generated by the model or obtained via syntactic mutations. For each training instance, we form a pair: $
\bigl((x, e_{1:t-1}, e_t^\star),\, (x, e_{1:t-1}, \tilde{e}_t)\bigr),$
with the preference label $e_t^\star \succ \tilde{e}_t$. StepPRM is trained using the standard preference-ranking objective (Bradley–Terry / logistic preference model) widely used in reward modeling and RLHF \cite{christiano2017deep,ouyang2022training}:
\[
\mathcal{L}_{\mathrm{PRM}}
=
- \mathbb{E}\! \left[
\log \sigma\!\left( 
V_{\phi}(x, e_{1:t-1}, e_t^\star)
-
V_{\phi}(x, e_{1:t-1}, \tilde{e}_t)
\right)
\right],
\]
where $\sigma(\cdot)$ is the sigmoid function. This objective encourages StepPRM to assign higher reward to semantically correct steps and penalize structurally invalid or logically incoherent ones.

\subsubsection{Reward Shaping via Partial Rollout Consistency}
Because RTL code is only verifiable when complete, StepPRM must infer step quality without explicit functional simulation. To address this, we introduce a consistency-based shaping term that leverages the structural alignment between partial implementations $c_t$ and the final canonical code $c^\star$. Let $A(c_t, c^\star)$ denote an alignment score computed via AST tree-edit similarity or structural matching. We define the shaped reward target for step $e_t$ as:
\[
y_t = \alpha \cdot \mathbf{1}[e_t \text{ from canonical}] + (1-\alpha) \cdot A(c_t, c^\star),
\]
where $\alpha \in [0,1]$ controls the balance between canonical supervision and structure-aware shaping. Intuitively, the indicator $\mathbf{1}[\cdot]$ provides a strong discrete signal when a step matches the canonical trace, while $A(\cdot,\cdot)$ supplies a continuous proxy for partial correctness when the step is novel or partially aligned. StepPRM is additionally trained to regress to this shaped reward:
\[
\mathcal{L}_{\mathrm{shaping}}
=
\mathbb{E}\! \left[
\bigl( V_{\phi}(x, e_{1:t}) - y_t \bigr)^2
\right],
\]
which aids calibration and provides denser gradients for generalization.

\subsubsection{Final Training Objective}
The full StepPRM loss combines preference learning and shaping:
\[
\mathcal{L}_{\mathrm{StepPRM}}
=
\mathcal{L}_{\mathrm{PRM}}
+
\lambda_{\mathrm{sh}} \, \mathcal{L}_{\mathrm{shaping}},
\]
where $\lambda_{\mathrm{sh}}$ weights the shaping term. This composite objective ensures that StepPRM captures both \emph{relative preference structure} between reasoning steps and \emph{absolute semantic quality} measured by partial structural consistency.

\subsubsection{Reward Assignment During Rollouts}
At inference or during MCTS-guided exploration, StepPRM assigns a reward: $r_t = V_{\phi}(x, e_{1:t})$
to each newly generated candidate step. These stepwise rewards provide dense, semantically grounded feedback, enabling efficient exploration and mitigating long-horizon credit assignment issues that commonly arise in RTL synthesis tasks.

\subsection{PRM-Guided MCTS}

To explore alternative reasoning paths beyond the canonical trajectories, we employ a PRM-guided Monte Carlo Tree Search (MCTS). In contrast to standard RLHF pipelines that apply rewards only after full-sequence generation, MCTS enables structured, branching exploration over partial RTL implementations. StepPRM provides dense, step-level evaluations that guide tree expansion, similar in spirit to value-guided planning in AlphaZero-style search \cite{wan2024alphazero} but adapted to the semantics of RTL synthesis.

\subsubsection{Search Tree Structure}
For a given specification $x$, MCTS constructs a search tree where each node represents a partial reasoning prefix $e_{1:t}$ and each edge corresponds to an operator-level or statement-level RTL step $e_{t+1}$. Each node maintains: $
\text{Node}(e_{1:t}) = \bigl(N_t,\, Q_t,\, \{a\},\, \{P_t(a)\}\bigr),$
where $N_t$ is the visit count, $Q_t$ is the accumulated step-value estimate, and $P_t(a)$ is the policy prior supplied by the current generation model $\pi_\theta(a \mid x, e_{1:t})$.

\subsubsection{StepPRM-Guided UCB Score}
During tree traversal, MCTS selects the next step by maximizing an Upper Confidence Bound (UCB) objective:
\[
a^\star 
= 
\arg\max_{a}
\left[
Q_t(a)
+
c_{\mathrm{uct}} \cdot P_t(a) \frac{\sqrt{\sum_b N_t(b)}}{1 + N_t(a)}
\right],
\]
where $c_{\mathrm{uct}}$ controls exploration.
Unlike typical MCTS where $Q_t(a)$ is backed up solely from terminal rewards, we initialize $Q_t(a) \leftarrow V_{\phi}(x, e_{1:t}, a)$,
directly using StepPRM.  
This provides dense, semantic feedback even for partial code, preventing uninformative plateaus common in long-horizon synthesis tasks.

\subsubsection{Rollout Expansion and PRM Value Backup}
When expanding a leaf node, the partial trajectory is extended using the policy model $\pi_\theta$ until a horizon depth $H$ or an early structural stopping criterion (e.g., balanced begin–end blocks) is met. StepPRM evaluates each new step, and the leaf value is computed as:
$
R_{\mathrm{leaf}} = \frac{1}{k} \sum_{i=1}^{k} V_{\phi}(x, e_{1:t+i}),
$
i.e., the average step-level semantic value across the rollout. This value is backed up through the tree:
\[
Q_t(a) \leftarrow 
\frac{N_t(a) \cdot Q_t(a) + R_{\mathrm{leaf}}}{N_t(a) + 1},
\quad
N_t(a) \leftarrow N_t(a)+1.
\]

\subsubsection{Balancing Exploration and Structural Feasibility}
To prevent exploration of syntactically invalid branches, MCTS performs feasibility checks on partial code $c_t$ (e.g., unmatched always-blocks, undeclared signals, combinational cycles).  
Branches that violate structural invariants are discarded and assigned a large negative StepPRM penalty via: $
V_{\phi}(x, e_{1:t}) \leftarrow -\beta,
$
where $\beta$ is a large constant.  
This tightens the search space and improves sample efficiency.

\subsubsection{Search Output and Trajectory Aggregation}
After $M$ simulations, the improved policy for each state is given by normalized visit counts: $
\hat{\pi}(a \mid x, e_{1:t})=\frac{N_t(a)^{\tau}}{\sum_b N_t(b)^{\tau}},
$
where $\tau$ is a temperature parameter.  
The top-ranked rollouts form an expanded dataset:
$
\mathcal{D}_{\mathrm{mcts}} = 
\{(x, \hat{\tau})\},
$
where $\hat{\tau}$ is a high-reward trajectory under StepPRM.  
These trajectories include novel, semantically consistent reasoning paths that go beyond canonical demonstrations, reducing bootstrap bias and stabilizing subsequent policy refinement (RAFT).

\subsection{Retrieval-Augmented Fine-Tuning (RAFT)}

After StepPRM-guided MCTS expands the trajectory space, the final component of StepPRM-RTL is Retrieval-Augmented Fine-Tuning (RAFT), which refines the generation policy $\pi_\theta$ using (i) retrieved repository-level context, and (ii) high-quality trajectories weighted by StepPRM-derived rewards. RAFT integrates retrieval-based grounding with reward-weighted policy optimization, enabling the policy to internalize both semantic reasoning structure and hardware-specific design patterns.

\subsubsection{Retrieval Model for Repository-Level Context}
Given a specification $x$, RAFT retrieves relevant RTL files, design patterns, module templates, or prior verified trajectories from a repository $\mathcal{R}$.  
We encode each repository element $d \in \mathcal{R}$ using a domain-tuned encoder $g(\cdot)$ and compute similarity with the query encoding $q(x)$:$
s(d \mid x) = \text{sim}(q(x), g(d))$
Top-$k$ documents, $\mathcal{C}(x) = \{ d_1, \dots, d_k \}$,
are retrieved \& concatenated with the trajectory prefix for conditioning.

\begin{table*}[h!]
\centering
\small
\begin{tabular}{l|cc|cc}
\toprule
\multirow{2}{*}{Model} & \multicolumn{2}{c|}{Pass@1} & \multicolumn{2}{c}{Reasoning Fidelity (\%)} \\
 & Verilog & VHDL & Verilog & VHDL \\
\midrule
\multicolumn{5}{c}{\textbf{Prompt-based Models}} \\
\midrule

Vanilla Prompting (GPT-4o) & 0.543 & 0.285 & - & - \\
CoDes (GPT-4o) & 0.602 & 0.348 & - & - \\\hline

\multicolumn{5}{c}{\textbf{Finetuning-based Models}} \\
\midrule

RTLCoder (Mistral) & 0.625 & - & - & - \\
CodeV (CodeQwen) & 0.532 & - & - & - \\
VeriThoughts & 0.755 & - & 60.4 & -\\\midrule
\multicolumn{5}{c}{\textbf{Finetuning-based Models}} \\
\midrule
RAG-CodeBERT (GPT-4o) & 0.688 & 0.487 & - & - \\
RAG-FT (GPT-4o) & 0.719 & 0.531 & -& - \\\hline
\multicolumn{5}{c}{\textbf{Our Model Variants}} \\
\hline
\textbf{StepPRM-RTL (Full)} & \textbf{0.857} & \textbf{0.786} & \textbf{82.5} & \textbf{80.2} \\
No MCTS (Sampling-Only) & 0.810 & 0.738 & 78.2 & 76.5 \\
Supervised RAFT Only & 0.796 & 0.721 & 75.3 & 73.0 \\
No PRM & 0.781 & 0.709 & 73.1 & 70.8 \\
\bottomrule
\end{tabular}
\caption{Overall performance of \textsc{StepPRM-RTL} compared to baselines. StepPRM-RTL achieves the highest Pass@1 and reasoning fidelity on both Verilog and VHDL benchmarks, demonstrating superior functional correctness and stepwise reasoning quality.}
\label{tab:overall_results}
\end{table*}

\subsubsection{Reward-Weighted Trajectories}
For each high-value MCTS trajectory $\hat{\tau} = \langle e_1, \dots, e_T\rangle$, StepPRM provides stepwise rewards: $r_t = V_{\phi}(x, e_{1:t}).$
We compute a normalized trajectory-level weight:
\[
w(\hat{\tau})
= 
\frac{\exp\!\left(\beta \sum_{t=1}^T r_t\right)}
{\sum_{\tau' \in \mathcal{D}} \exp\!\left(\beta \sum_{t} r'_t\right)},
\]
where $\beta$ controls reward sensitivity.  
This weighting mechanism resembles advantage-weighted regression in RL fine-tuning and preference-based policy optimization, but adapted to step-level rewards and trajectory supervision.

\subsubsection{Policy Update}
The policy is fine-tuned to maximize the likelihood of high-value trajectories given the retrieved context:
\[
\mathcal{L}_{\mathrm{RAFT}}
=
- \mathbb{E}_{(x, \hat{\tau}) \sim \mathcal{D}_{\mathrm{mcts}}}
\left[
w(\hat{\tau}) 
\sum_{t=1}^T 
\log \pi_\theta(e_t \mid x, \mathcal{C}(x), e_{1:t-1})
\right].
\]
This objective encourages the model to reproduce high-reward decision sequences while grounding them in repository-level retrieved signals.  Compared to standard supervised fine-tuning, RAFT introduces two key improvements: (a) retrieval grounding, which exposes the policy to reusable structural patterns and contextually relevant design idioms drawn from existing repositories; and (b) reward weighting, which prioritizes trajectories that StepPRM and MCTS jointly deem semantically consistent and structurally valid.
% Compared to standard supervised fine-tuning, RAFT introduces two critical enhancements: (a) \textbf{Retrieval grounding} ensures the policy learns reusable structural patterns and contextually relevant idioms from existing design repositories; and (b) \textbf{Reward weighting} ensures the model preferentially internalizes reasoning trajectories that StepPRM and MCTS identified as semantically consistent and structurally valid.

\subsubsection{Iterative Integration with StepPRM and MCTS}
RAFT closes the StepPRM-RTL loop. After each RAFT update, which is:
$\pi_\theta^{(k+1)} \leftarrow \text{RAFT}\bigl(\pi_\theta^{(k)}\bigr)$, the improved policy becomes the proposal distribution for the next iteration. This yields higher-quality search rollouts, which in turn allow StepPRM to refine reward estimates from a more diverse and semantically richer trajectory distribution.

% The Step-Level Process Reward Model (StepPRM) is a transformer that predicts scalar rewards per step conditioned on the partial trajectory and specification.
\subsection{Implementation Details}
We implement \textsc{StepPRM-RTL} in PyTorch~\cite{imambi2021pytorch}, fine-tuning Qwen3-8B-Instruct~\cite{yang2025qwen3} on stepwise trajectories using 2--4 NVIDIA A100 GPUs. StepPRM takes as input the concatenation of the spec, partial code state, and current step (rationale + edit), encoded via a transformer with a scalar regression head. Retrieval uses Qwen3-Embedding-4B~\cite{zhang2025qwen3} trained via contrastive learning on HDL repositories, with top-$k$ matches prepended during RAFT fine-tuning. Structured exploration is performed via MCTS-guided by StepPRM, using 50 simulations per specification, an exploration constant $c_{\mathrm{uct}} = 1.5$, and a rollout horizon of 10 steps. StepPRM rewards are combined with a structural alignment term ($\lambda_{\mathrm{sh}}=0.5$) for reward shaping. Outcome verification employs Icarus Verilog for Verilog and GHDL+VUnit for VHDL, though StepPRM scores primarily guide MCTS. The training pipeline first pretrains the policy and StepPRM on canonical trajectories, expands the trajectory space via StepPRM-guided MCTS, and refines the policy with reward-weighted RAFT using retrieved context, iteratively improving both policy and reward model.

\section{Experiments}

We evaluate \textsc{StepPRM-RTL} on RTL synthesis using two benchmarks: Verilog-Eval~\cite{verilogeval} (156 spec-to-Verilog tasks with self-checking testbenches from HDLBits) and VHDL-Eval~\cite{vijayaraghavan2024vhdl} (202 translated VHDL tasks with similar verification). We augment both with LLM-generated, stepwise rationales validated via intermediate checks, enabling evaluation of correctness and reasoning. We compare against finetuned baselines—\textbf{VeriThoughts}~\cite{yubeaton2025verithoughts}, \textbf{Verigen}~\cite{verigen}, \textbf{RTLCoder}~\cite{liu2024rtlcoder}, \textbf{CodeV~\cite{codev}} and the VHDL baseline \textbf{CoDes}~\cite{vijayaraghavan2024chain}.
%We empirically evaluate \textsc{StepPRM-RTL} on RTL synthesis using two benchmark suites: the Verilog-Eval~\cite{verilogeval} benchmark (156 specification-to-Verilog tasks with built-in self-verifying testbenches derived from HDLBits) and the more recent VHDL-Eval~\cite{vijayaraghavan2024vhdl} benchmark (202 VHDL tasks translated from Verilog-Eval and public tutorials, using analogous self-verification structures). These benchmarks are further augmented with stepwise rationales generated using LLMs and verified through intermediate reasoning steps, providing high-quality trajectories for evaluating both functional correctness and reasoning fidelity. We compare our method against competitive Verilog-oriented finetuned baselines including \textbf{VeriThoughts}~\cite{yubeaton2025verithoughts}, \textbf{Verigen}~\cite{verigen}, \textbf{RTLCoder}~\cite{liu2024rtlcoder}, and \textbf{CodeV}~\cite{codev}, as well as one of the VHDL baseline \textbf{CoDes}~\cite{vijayaraghavan2024chain}. 
We also evaluate strong RAG-enabled LLM baselines: \textbf{GPT-4o}~\cite{chatgpt} and \textbf{Qwen3-8B} (RAG, no PRM/MCTS)~\cite{bai2023qwen}. ll models are evaluated on two primary metrics: \emph{Pass@1}, computed using the official testbenches, and \emph{reasoning fidelity}, measured by an LLM judge comparing generated reasoning trajectories against canonical reasoning steps in each benchmark. Our experiments address two research questions:
\textbf{RQ1:} How does \textsc{StepPRM-RTL} compare to state-of-the-art baselines on Verilog/VHDL synthesis?
\textbf{RQ2:} What is the contribution of each pipeline component?
% We additionally evaluate strong retrieval-augmented large-model baselines such as \textbf{GPT-4o}~\cite{chatgpt} and a RAG-enabled \textbf{Qwen3-8B} (without PRM or MCTS)~\cite{bai2023qwen}. All models are evaluated on two primary metrics: \emph{Pass@1}, computed using the official testbenches, and \emph{reasoning fidelity}, measured by an LLM judge comparing generated reasoning trajectories against canonical reasoning steps in each benchmark. Our experiments address two research questions: \textbf{RQ1:} How does \textsc{StepPRM-RTL} compare to state-of-the-art baselines across Verilog and VHDL synthesis tasks? 
% \textbf{RQ2:} How do the components of the pipeline contribute to overall performance?

\section{Results}

% We evaluate \textsc{StepPRM-RTL} on RTL synthesis using two benchmark suites, Verilog-Eval~\cite{} and VHDL-Eval~\cite{}. Verilog-Eval comprises 156 specification-to-Verilog tasks with built-in testbenches, derived from HDLBits, while VHDL-Eval contains 202 VHDL tasks translated from Verilog-Eval and public tutorials with analogous self-verification structures. These benchmarks are further augmented with step-wise rationales generated using LLMs and verified using intermediate steps. We compare against a range of baselines: prompt-based models (\textbf{Vanilla Prompting} GPT-4o, \textbf{CoDes} GPT-4o), finetuning-based models (\textbf{RTLCoder} Mistral, \textbf{CodeV} CodeQwen, \textbf{VeriThoughts}), RAG-enabled LLMs (\textbf{RAG-CodeBERT}, \textbf{RAG-FT} GPT-4o), and our full StepPRM-RTL framework.  

\subsection{RQ1: Overall Results}
\textsc{StepPRM-RTL} achieves the highest Pass@1 and reasoning fidelity on both Verilog and VHDL benchmarks, as shown in Table~\ref{tab:overall_results}. Pass@1 measures the probability that the first generated implementation passes the functional testbench, while reasoning fidelity quantifies how closely the model's intermediate rationales align with ground-truth stepwise reasoning in the benchmark. StepPRM-RTL consistently outperforms prompt-based and finetuning-based baselines, with 0.857 and 0.786 Pass@1 on Verilog and VHDL, respectively, and reasoning fidelity exceeding 80\%. Compared to RAG-FT, StepPRM-RTL leverages dense StepPRM rewards and MCTS exploration to improve intermediate reasoning and final correctness. Prompt-based models (Vanilla, CoDes) lag due to no trajectory supervision, while finetuned models show moderate gains, underscoring the value of reward-guided trajectory learning.
%Compared to RAG-FT, StepPRM-RTL benefits from dense StepPRM rewards and MCTS-guided exploration, which improve intermediate reasoning and final functional correctness. Prompt-based models (Vanilla, CoDes) lag behind due to the absence of trajectory supervision, while finetuning-based models show moderate gains, highlighting the importance of reward-guided trajectory learning.

\subsection{RQ2: Ablation Studies}

To quantify the contribution of each StepPRM-RTL component, we conduct ablation experiments on MCTS, the StepPRM, and reward-based RAFT. Results are reported in Table~\ref{tab:overall_results}.

\noindent \textbf{Impact of MCTS Search}
We disable PRM-guided MCTS and replace it with sampling-only rollouts, generating $K=20$ candidate trajectories per specification. These trajectories are fed into RAFT fine-tuning, retaining StepPRM-based reward weighting. Without structured MCTS, Pass@1 decreases from 0.857 to 0.810 on Verilog and 0.786 to 0.738 on VHDL ($\approx$4.7--5.0 pp drop), while reasoning fidelity drops by 4--4.5 pp. This confirms that MCTS is critical for selecting high-quality intermediate steps, reducing invalid rollouts, and effectively exploring diverse reasoning paths that naive sampling cannot cover.  

\noindent \textbf{Outcome vs. Process Rewards}
To isolate the effect of the StepPRM, we remove step-level reward supervision and rely solely on outcome-based verification (i.e., functional correctness checked via Icarus Verilog for Verilog, GHDL/VUnit for VHDL). The same MCTS and RAFT pipeline is retained. Removing PRM leads to Pass@1 dropping from 0.857 to 0.781 ($\approx$7.6 pp) on Verilog and 0.786 to 0.709 on VHDL, with reasoning fidelity falling from 82.5\% to 73.1\% and 80.2\% to 70.8\%, respectively. These results indicate that outcome-only rewards, even with simulator/formal verification, provide sparse feedback insufficient for guiding intermediate step-level reasoning. StepPRM supplies dense, interpretable rewards, improving both long-horizon reasoning and trajectory quality.

\noindent \textbf{Influence of Reward-Based RAFT (Supervised-Only RAFT)}
We also evaluate RAFT fine-tuning without reward weighting, i.e., treating all high-value trajectories equally regardless of StepPRM scores. In this configuration, Pass@1 drops from 0.857 to 0.796 on Verilog and 0.786 to 0.721 on VHDL ($\approx$6 pp), while reasoning fidelity decreases by 7--8 pp. This demonstrates that reward-guided RAFT is necessary to prioritize semantically high-quality steps and not just reproduce trajectory sequences. Thus, these ablation studies demonstrate that each StepPRM-RTL component is crucial: MCTS enables structured exploration and reduces invalid rollouts, PRM provides dense step-level rewards for intermediate reasoning, and reward-weighted RAFT consolidates high-quality trajectories into the policy. Removing any component degrades both functional correctness and reasoning fidelity, validating the design decisions of our framework.
\subsection{Hyperparameter Sensitivity Analysis}

\begin{figure}[h]
    \centering
    \includegraphics[width=\linewidth]{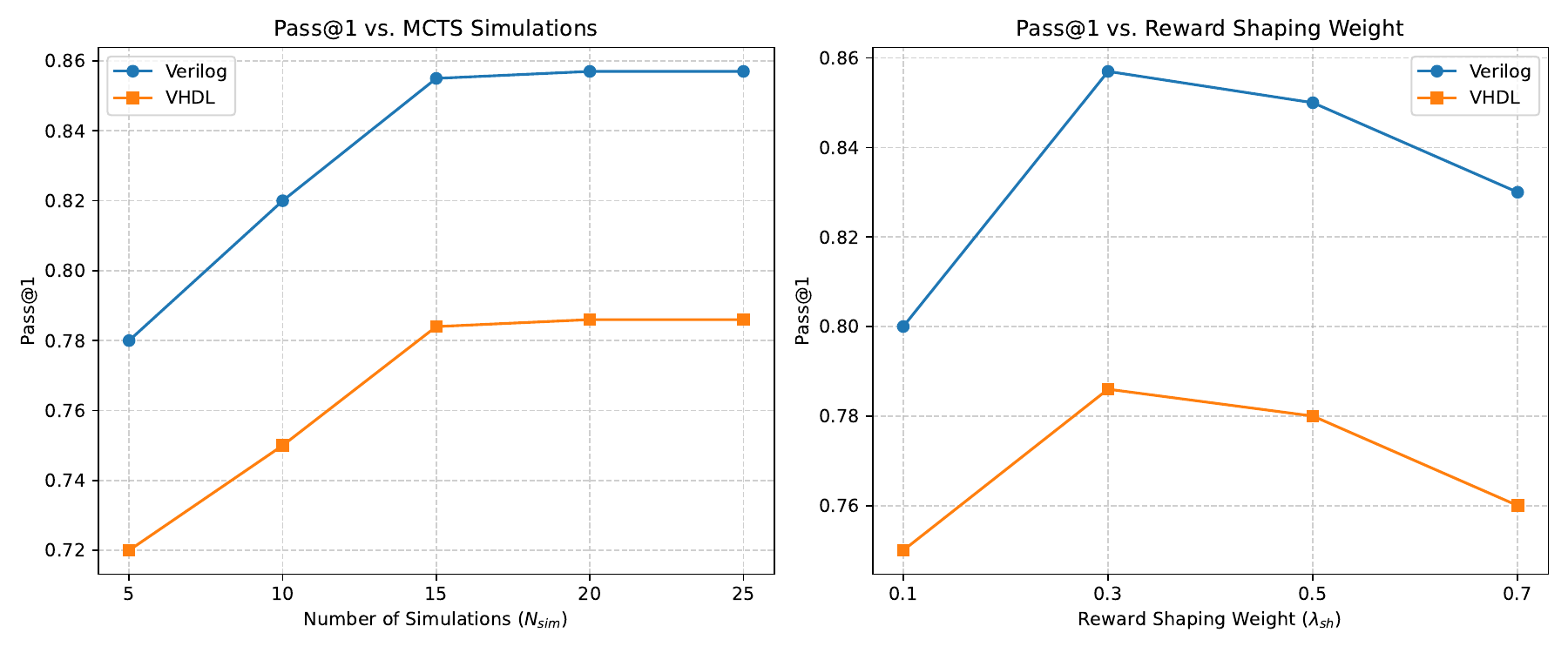}
    \caption{Hyperparameter sensitivity for StepPRM-RTL. Left: Pass@1 vs. MCTS simulations ($N_\text{sim}$). Right: Pass@1 vs. shaping weight ($\lambda_\text{sh}$). Best performance at $N_\text{sim}=15$ and $\lambda_\text{sh}=0.3$.}
    % Hyperparameter sensitivity analysis for StepPRM-RTL. \textbf{Left:} Pass@1 versus number of MCTS simulations ($N_\text{sim}$). \textbf{Right:} Pass@1 versus reward shaping weight ($\lambda_\text{sh}$). Optimal tradeoffs are observed at $N_\text{sim}=15$ and $\lambda_\text{sh}=0.3$.}
    \label{fig:hyperparam_sensitivity}
\end{figure}

We analyze the impact of two critical hyperparameters on StepPRM-RTL performance: the number of MCTS simulations per specification ($N_\text{sim}$) and the reward shaping weight ($\lambda_\text{sh}$). Figure~\ref{fig:hyperparam_sensitivity} shows Pass@1 results on Verilog and VHDL benchmarks for both hyperparameters. \textbf{MCTS Simulation Count:} Increasing $N_\text{sim}$ improves Pass@1, rising from $0.78$ to $0.857$ for Verilog and $0.72$ to $0.786$ for VHDL as simulations increase from $5$ to $25$. Notably, $N_\text{sim}=15$ achieves nearly the same performance as $20$–$25$ simulations, offering a favorable tradeoff between accuracy and computational cost. Gains plateau beyond $15$ simulations, suggesting that StepPRM effectively prioritizes high-value steps. \textbf{Reward Shaping Weight:} Pass@1 peaks at $\lambda_\text{sh}=0.3$, striking a balance between canonical step preference and structural alignment. Lower values underweight structural guidance, while higher values ($\lambda_\text{sh} \ge 0.5$) overemphasize alignment, occasionally penalizing creative yet correct steps. The trend is consistent across Verilog and VHDL benchmarks, confirming robustness across architectures. Overall, StepPRM-RTL performance is stable for moderate MCTS simulations and shaping weights, providing a practical tradeoff between exploration, step-level guidance, and computational cost.

% \subsection{Hyperparameter Sensitivity Analysis}

% We analyze the impact of two critical hyperparameters on StepPRM-RTL performance: the number of MCTS simulations per specification ($N_\text{sim}$) and the reward shaping weight ($\lambda_\text{sh}$). Pass@1 results on Verilog and VHDL benchmarks are shown in Figure~\ref{fig:hyperparam_sensitivity}.

% \textbf{MCTS Simulation Count:} Increasing $N_\text{sim}$ improves Pass@1, rising from $0.78$ to $0.857$ for Verilog and $0.72$ to $0.786$ for VHDL as simulations increase from $5$ to $25$. Gains plateau beyond $20$ simulations, suggesting that StepPRM effectively prioritizes high-value steps and additional rollouts yield diminishing returns.

% \textbf{Reward Shaping Weight:} Pass@1 peaks at $\lambda_\text{sh}=0.5$, balancing canonical step preference and structural alignment. Lower values underweight structural guidance, while higher values overemphasize alignment, occasionally penalizing creative but correct steps. Verilog and VHDL results consistently reflect this trend, indicating robustness across architectures.

% These results confirm that StepPRM-RTL performance is stable for moderate MCTS simulation counts and shaping weights, providing a practical tradeoff between exploration, step-level guidance, and computational cost.

\section{Conclusion}

We introduced \texttt{StepPRM-RTL}, an RL-guided framework for RTL synthesis that integrates stepwise trajectory decomposition, a Step-Level Process Reward Model (StepPRM), PRM-guided MCTS, and retrieval-augmented fine-tuning (RAFT). By modeling RTL generation as a sequence of semantically meaningful steps with dense, interpretable rewards, StepPRM-RTL effectively addresses long-horizon credit assignment and ensures both intermediate reasoning fidelity and final functional correctness. Experiments on Verilog-Eval and VHDL-Eval benchmarks show that StepPRM-RTL outperforms prompt-based, fine-tuned, and retrieval-augmented LLM baselines, achieving state-of-the-art Pass@1 and reasoning fidelity. Ablation studies highlight the critical role of structured MCTS exploration, step-level rewards, and reward-weighted RAFT in improving trajectory quality and reasoning. Future directions include extending the framework to multi-file hierarchical designs, integrating formal verification more tightly into the reward model, and exploring cross-architecture transfer of reasoning trajectories. Overall, StepPRM-RTL bridges interpretable stepwise reasoning with improved RTL synthesis, providing a promising foundation for AI-assisted hardware design.

\bibliographystyle{ACM-Reference-Format}
\bibliography{main}

% %%
% %% If your work has an appendix, this is the place to put it.
% \appendix









\end{document}